\documentclass[acmsmall, screen]{acmart}

\usepackage{algorithmic}

\usepackage{graphicx}
\usepackage{multirow}
\usepackage{booktabs}
\usepackage{subfigure}
\usepackage{amsmath,amsfonts,bm}
\usepackage{float}
\usepackage{enumitem}
\usepackage[normalem]{ulem}
\useunder{\uline}{\ul}{}

\usepackage[linesnumbered,ruled,vlined]{algorithm2e}

\newtheorem{definition}{Definition}
\newtheorem{problem}{Problem}

\AtBeginDocument{%
  }

\setcopyright{acmlicensed}
\copyrightyear{2018}
\acmYear{2018}
\acmDOI{XXXXXXX.XXXXXXX}

\acmJournal{JACM}
\acmVolume{37}
\acmNumber{4}
\acmArticle{111}
\acmMonth{8}

\begin{document}

\title{MagiNet: Mask-Aware Graph Imputation Network for Incomplete Traffic Data}


\author{Jianping Zhou}
\email{jianpingzhou@sjtu.edu.cn}
\affiliation{%
  \institution{Shanghai Jiao Tong University}
  \city{Shanghai}
  \country{China}
}

\author{Bin Lu}
\email{robinlu1209@sjtu.edu.cn
}
\affiliation{%
  \institution{Shanghai Jiao Tong University}
  \city{Shanghai}
  \country{China}
}

\author{Zhanyu Liu}
\email{zhyliu00@sjtu.edu.cn
}
\affiliation{%
  \institution{Shanghai Jiao Tong University}
  \city{Shanghai}
  \country{China}
}

\author{Siyu Pan}
\email{pansiyu0327@sjtu.edu.cn
}
\affiliation{%
  \institution{Shanghai Jiao Tong University}
  \city{Shanghai}
  \country{China}
}

\author{Xuejun Feng}
\email{fengxuejun@sjtu.edu.cn
}
\affiliation{%
  \institution{Shanghai Jiao Tong University}
  \city{Shanghai}
  \country{China}
}

\author{Hua Wei}
\email{hua.wei@asu.edu
}
\affiliation{%
  \institution{Arizona State University}
  \city{Arizona}
  \country{USA}
}

\author{Guanjie Zheng}
\affiliation{%
  \institution{Shanghai Jiao Tong University}
  \city{Shanghai}
  \country{China}}
\email{gjzheng@sjtu.edu.cn}

\author{Xinbing Wang}
\affiliation{%
 \institution{Shanghai Jiao Tong University}
 \city{Shanghai}
 \country{China}
 }
\email{xwang8@sjtu.edu.cn}

\author{Chenghu Zhou}
\affiliation{%
  \institution{Chinese Academy of Sciences}
  \city{Beijing}
  \country{China}
  }
\email{zhouchsjtu@gmail.com}
\renewcommand{\shortauthors}{Jianping Zhou et al.}


\received{20 February 2007}
\received[revised]{12 March 2009}
\received[accepted]{5 June 2009}

\begin{abstract}
Due to detector malfunctions and communication failures, missing data is ubiquitous during the collection of traffic data.
Therefore, it is of vital importance to impute the missing values to facilitate data analysis and decision-making for Intelligent Transportation System (ITS).
However, existing imputation methods generally perform \emph{zero} pre-filling techniques to initialize missing values, introducing inevitable noises.
Moreover, we observe prevalent over-smoothing interpolations, falling short in revealing the intrinsic spatio-temporal correlations of incomplete traffic data.
To this end, we propose \emph{\underline{M}ask-\underline{a}ware \underline{g}raph \underline{i}mputation \underline{Net}work}: \textbf{MagiNet}.
Our method designs an adaptive mask spatio-temporal encoder to learn the latent representations of incomplete data, eliminating the reliance on pre-filling missing values.
Furthermore, we devise a spatio-temporal decoder that stacks multiple blocks to capture the inherent spatial and temporal dependencies within incomplete traffic data, alleviating over-smoothing imputation.
Extensive experiments demonstrate that our method outperforms state-of-the-art imputation methods on five real-world traffic datasets, yielding an average improvement of 4.31\% in RMSE and 3.72\% in MAPE.
\end{abstract}

\maketitle

\section{Introduction}
Missing data problems are pervasive when collecting traffic data from the Intelligent Transportation System (ITS)~\cite{zheng2016urban,zhang2020curb,xu2023uncovering,wang2017large}, owing to numerous uncontrollable factors, such as detector malfunctions, communication failures, and maintenance issues~\cite{gong2021missing,qin2021network,wang2023gstae,saha2019imputing}.
Incomplete traffic data hinders comprehensive data analysis and decision-making.
Consequently, how to impute the missing values in traffic data arises as an important problem.

\begin{figure}[t]
    \centering
    \includegraphics[width=\textwidth]{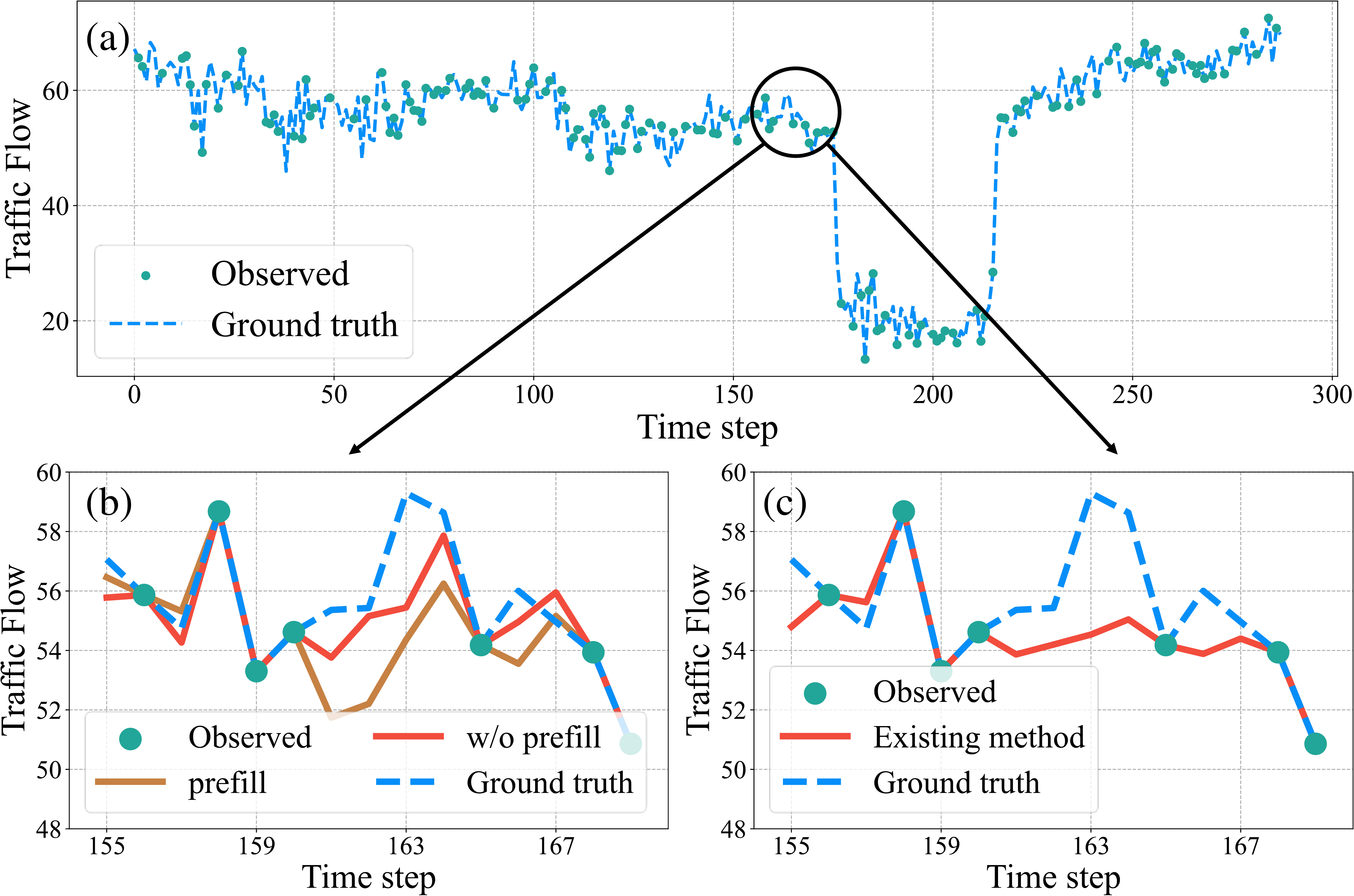}
    \caption{(a) Example of incomplete traffic data and corresponding ground truth from the Seattle dataset. (b) Performance comparison with and without pre-filling techniques to initialize missing values. Imputation performance with pre-filling techniques is significantly worse than without pre-filling. (c) Existing methods fail to capture inherent spatio-temporal correlations in incomplete traffic data, leading to an over-smoothing effect at dynamic missing positions, particularly noticeable between time steps 160 and 168.}
    \label{fig:introduction}
\end{figure}

The first and foremost issue lies in the initialization of missing values.
Specifically, how to replace the missing value placeholder, i.e., \emph{NaN}, for feature extraction.
Traditional statistical methods~\cite{song2015knn,kong2013matrixfactorization,white2011mice} do not initialize missing values. Instead, they directly impute missing values utilizing the statistical indicators of available observations.
These approaches rely on the strong assumption of time series stationarity, which fails to capture the nonlinear properties in incomplete traffic data, leading to poor performance. 
Recent deep imputation methods~\cite{cao2018brits,yoon2018gain,andrea2022grin,xu2022gagan} primarily employ \emph{zero pre-filling} to initialize missing values and use a mask matrix to record their positions.
The incomplete data is then treated as \emph{complete} data to ensure spatio-temporal feature learning.
However, utilizing pre-filling techniques to initialize missing values inevitably introduces noise and misleads the feature learning process.
We present an example of incomplete traffic data from the Seattle dataset for a single day in Figure~\ref{fig:introduction}(a).
Subsequently, we compare the imputation performance with and without pre-filling missing values.
As shown in Figure~\ref{fig:introduction}(b), the imputation performance with pre-filling missing values is significantly worse than without pre-filling.

Another issue is capturing the spatial and temporal correlations within incomplete traffic data.
To be specific, how to aggregate observation information to impute missing values while avoiding the over-smoothing effect.
Many traffic forecasting methods~\cite{yu2017stgcn,li2017dcrnn,wu2019gwnet,lan2022dstagnn,shao2022d2stgnn} have conducted research on capturing spatio-temporal dependencies, but they rely on complete history data, rendering them unsuitable for missing scenarios.
Moreover, most imputation methods~\cite{cao2018brits,yoon2018gain,richardson2020mcflow} do not simultaneously consider the complex spatial and temporal dependencies. 
Recent deep models tailored for spatio-temporal imputation, such as  GRIN~\cite{andrea2022grin} and GA-GAN~\cite{xu2022gagan}, utilize pre-filled data to capture spatio-temporal correlations.
However, these approaches ignore inherent dynamic changes, resulting in an over-smoothing effect, as shown in Figure~\ref{fig:introduction}(c).

Therefore, to address the above two issues, we propose \textit{\underline{M}ask-\underline{a}ware \underline{g}raph \underline{i}mputation \underline{Net}work} in an encoder-decoder framework, named \textbf{MagiNet}.
The core idea of our method is to adaptively learn the representation of missing values and capture intrinsic spatio-temporal correlations within incomplete data.
Firstly, we propose an adaptive mask spatio-temporal encoder (\emph{AMSTenc}) that generates representations of missing values through a learnable missing encoding, which adaptively learns inherent representations without introducing supplementary noise.
Additionally, we utilize a mask-aware spatio-temporal decoder (\emph{MASTdec}) that stacks multiple blocks, adopting a mask-aware attention mechanism.
The decoder incorporates the adaptive missing embedding to consciously adjust the aggregation coefficient with observation embedding, extracting inherent correlations within incomplete traffic data and alleviating the over-smoothing effect.

In summary, the main contributions are as follows:
\begin{itemize}[leftmargin=*]
\item We design an adaptive mask spatio-temporal encoder that does not require pre-filling techniques. It dynamically learns the representation of incomplete traffic data, effectively eliminating the introduction of uncontrollable noise.
\item We design a spatio-temporal decoder that incorporates a mask-aware attention mechanism to capture inherent dependencies within incomplete data and alleviate the over-smoothing effect.
\item Experimental results on five real-world traffic datasets show that our method outperforms state-of-the-art imputation methods, with an average improvement of 4.31\% in RMSE and 3.72\% in MAPE, and consistently exhibits improvement across different missing ratios.
\end{itemize}
\section{Related Work}
\subsection{Missing Data Imputation}
Existing methods for missing data imputation could be generally divided into three categories: \textit{statistical learning-based methods}, \textit{machine learning-based methods}, and \textit{deep learning-based methods}.
(1) The \textit{statistical learning-based methods}, such as Mean, Median, KNN~\cite{song2015knn}, and MICE~\cite{white2011mice}, impute missing values based on the statistical properties of the available observations, assuming the time series is stationary.
However, they fail to capture the nonlinear characteristics inherent in incomplete traffic data.
(2) The \textit{machine learning-based methods} include the traditional probabilistic principal component analysis~\cite{qu2009ppca}, matrix factorization~\cite{fan2022dynamic,he2022bayesian,jia2021missing} and EM algorithms~\cite{mirzaei2022missing}.
However, these methods cannot simultaneously learn the dynamic spatio-temporal correlations, and the computational complexity increases significantly when handling high-dimensional data.
(3) Recently, a plethora of \textit{deep learning-based methods} have been proposed. 
BRITS~\cite{cao2018brits} utilizes the bidirectional recurrent neural network architecture to capture temporal features.
MIWAE~\cite{mattei2019miwae}, GAIN~\cite{yoon2018gain}, MISGAN~\cite{li2019misgan}, Mcflow~\cite{richardson2020mcflow}, 
ST-SCL~\cite{stscl} treat the imputation task as the temporal generation task, incorporating popular generative architectures, such as AutoEncoder~\cite{kingma2013auto}, GAN~\cite{goodfellow2020gan}, and Flow~\cite{kingma2018glow}.
However, traffic data is a type of data with temporal and spatial characteristics, and obviously, these methods ignore the modeling of spatial features.
GRIN~\cite{andrea2022grin}, GA-GAN~\cite{xu2022gagan} involve graph neural networks to capture spatial correlations, but these methods utilize pre-filling techniques to initialize missing values, introducing supplementary noises. 
PriSTI~\cite{liu2023pristi} employs MPNN and diffusion modules to extract features, but its message-passing mechanism falls short in capturing inherent spatio-temporal correlation within incomplete traffic data. 

\subsection{Modeling the Spatio-temporal Dependency}
In recent years, numerous deep models tailored for spatio-temporal data have been developed to capture spatial and temporal dependencies, such as STGCN~\cite{yu2017stgcn}, DCRNN~\cite{li2017dcrnn}, Graph WaveNet~\cite{wu2019gwnet}, and others~\cite{lan2022dstagnn,shao2022d2stgnn,yi2019citytraffic,han2021dynamic,wang2020traffic,cao2022spatio}. 
These methods design diverse graph convolution modules to extract spatio-temporal features for traffic forecasting task, where the primary objective is to predict future traffic data by leveraging complete history data. 
However, in the real world, the missing data problem is pervasive in traffic data, preventing these methods from accessing complete traffic data to capture spatio-temporal dependencies.
In this paper, our task concentrates on imputing missing values based on observations from the incomplete data, rather than utilizing complete history data to forecast future data.
Therefore, we need to address how to capture inherent spatio-temporal dependencies within incomplete traffic data instead of complete data.

\section{Preliminary}
In this section, we first define the concept of incomplete traffic data and then give the problem definition of imputation task.

\begin{definition}
    Due to real-world detectors commonly manifest a non-uniform spatial distribution, we define the incomplete traffic data as graph-based data $\mathcal{D}=\{\mathcal{G},(\mathbf{X}_t,\mathbf{M}_t)_{t=1}^T\}$, spanning $T$ time steps.
$\mathcal{G}$ denotes the underlying traffic network, actually $\mathcal{G}=(\mathcal{V},\mathcal{E},\textbf{A})$, where $\mathcal{V}$ is a finite set of $|\mathcal{V}|=N$ nodes, $\mathcal{E}$ is a set of connecting edges between nodes in $\mathcal{V}$, and $\mathbf{A}$ is the adjacency matrix of $\mathcal{G}$. 
$\mathbf{X}_t=(\mathbf{x}_t^1,\mathbf{x}_t^2,\cdots,\mathbf{x}_t^N)\in \mathbb{R}^{N\times C}$ denotes the incomplete values of all nodes at the timestamp $t$, where $\mathbf{x}_t^i \in \mathbb{R}^{C}$ denotes the incomplete values of $C$ features of node $i$ at the timestamp $t$.  
$\mathbf{M}_t=(m_t^1,m_t^2,\cdots,m_t^N)\in \{0,1\}^{N}$ is the mask matrix.
$m_t^i\in \{0,1\}$ denotes whether the $\mathbf{x}^i_t$ is observed: 
$1$ indicates $\mathbf{x}^i_t$ is observed, and $0$ indicates $\mathbf{x}^i_t$ is missing. 
\end{definition}

\begin{problem}
    \textbf{Incomplete Traffic Data Imputation.}
    Suppose we have incomplete traffic data over $\tau$ time slices, and we want to impute the missing values. 
    The imputation task is formulated as learning a function $f(\cdot)$ given a underlying graph $\mathcal{G}$:
\begin{equation}
    [(\mathbf{X}_1,\mathbf{M}_1),\cdots,(\mathbf{X}_\tau,\mathbf{M}_\tau);\mathcal{G}]\stackrel{f(\cdot)}{\longrightarrow}[\hat{\mathbf{X}}_1,\cdots,\hat{\mathbf{X}}_\tau],
\end{equation}
such that minimizes the estimation error on missing positions, where $\hat{\mathbf{X}}_t\in \mathbb{R}^{N\times C}$ denotes the imputed data at the timestamp $t$.
\end{problem}
In this paper, all the notation is summarized in Table~\ref{tab:notation} for brevity.

\begin{table}[htbp]
\centering
\caption{Notations.}
\label{tab:notation}
\resizebox{\textwidth}{!}{%
\begin{tabular}{@{}c|c|c|c@{}}
\toprule
Notation & Definition & Notation & Definition \\ \midrule
   $\mathcal{D}$    &  incomplete traffic dataset       & $\mathbf{X}_o$ & the observation embedding        \\
   $\mathcal{G}$    &   the underlying traffic graph   &$\mathbf{Z}_u$ & the learnable missing embedding   \\ 
   $\mathcal{V}$    &  the node set of $\mathcal{G}$ & $\mathbf{X}_p$ &  the incomplete traffic data embedding  \\
   $\mathcal{E}$    &  the edge set of $\mathcal{G}$  & $\mathbf{H}$ &  the latent representation after \textit{AMSTenc}     \\
   $\textbf{A}$    &  the adjacency matrix of $\mathcal{G}$    & $\mathbf{T}_{att}$ &  the mask-aware temporal attention score     \\
   $\mathbf{X}_t$    &   the incomplete traffic data at timestamp $t$  & $\mathbf{A}^{(l)}$ & the temporal attention score of the $l$-th block       \\ 
   $\mathbf{M}_t$    &  the mask matrix of $\mathbf{X}_t$  & $\mathbf{H}_{matt}$ &  the representation after temporal attention aggregation      \\ 
   $\hat{\mathbf{X}}_t$    &  the imputed traffic data   & $\mathbf{H}^{\prime}_{matt}$  & the representation after spatial attention aggregation      \\ 
   $f(\cdot)$    &  the function to impute the missing values    & $\mathbf{S}_{att}$ & the spatial attention score    \\ 
   $N$    &  the number of nodes     & $\mathbf{E}$ & the representation after graph convolution    \\ 
   $T$    &  the length of time steps     & $\mathbf{E}_{out}$ & the representation after multi-scale temporal convolution    \\ 
   $C$    &  the number of features     & $\mathbf{H}_{out}$ & the output of mask-aware spatio-temporal block   \\ 
   \bottomrule
\end{tabular}%
}
\end{table}
\begin{figure}[!t]
    \centering    
    \includegraphics[width=\textwidth]{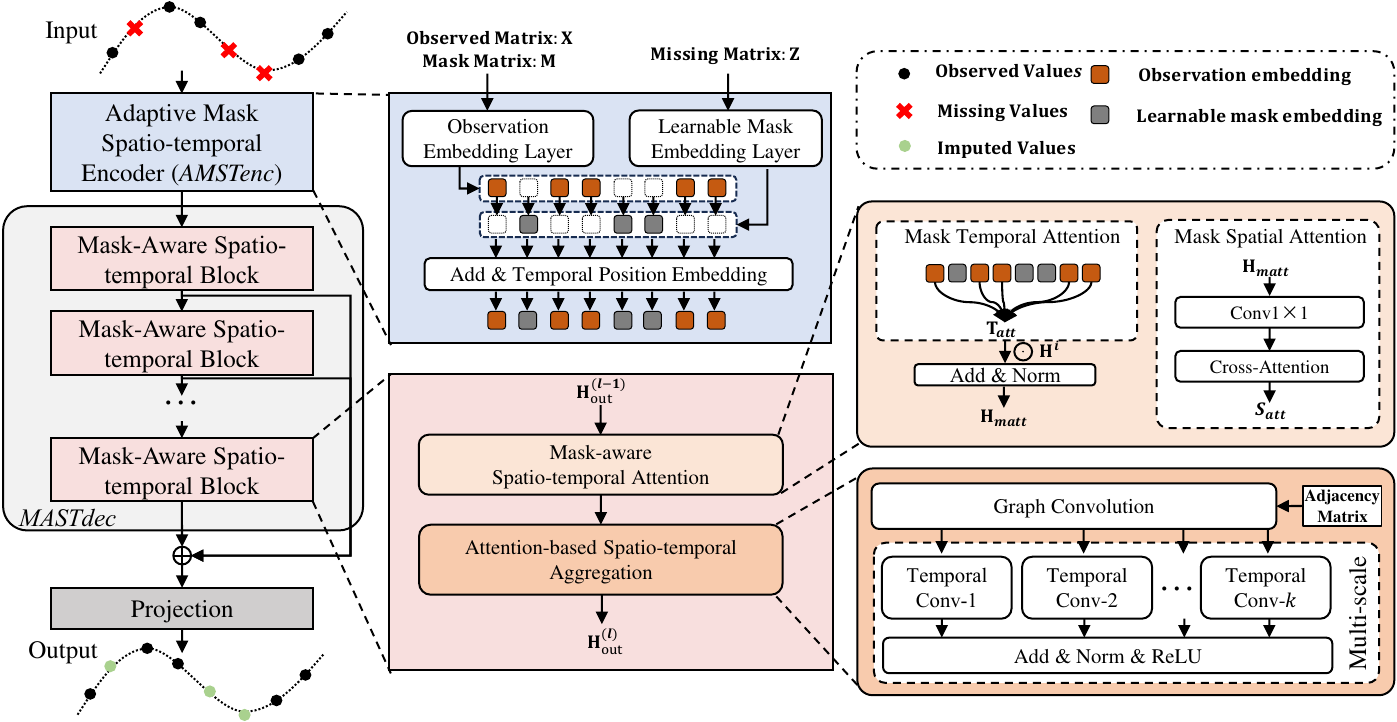}
    \caption{An overview of MagiNet, which consists of an adaptive mask spatio-temporal encoder (\textit{AMSTenc}) and a mask-aware spatio-temporal decoder (\textit{MASTdec}). The \textit{MASTdec} stacks several spatio-temporal blocks (\textit{ST Block}). 
    Each \textit{ST Block} combines a mask-aware spatio-temporal attention (\textit{MASTatt}) module that calculates the mask temporal and spatial attention, an attention-based spatio-temporal aggregation module that aggregates spatial information using graph convolution and multi-scale temporal information using temporal convolution.}
    \label{fig:MagiNet}
\end{figure}

\section{Methodology}
In this section, we introduce our model, MagiNet, to tackle the incomplete traffic data imputation task.
The proposed MagiNet, shown in Figure~\ref{fig:MagiNet}, includes an adaptive mask spatio-temporal encoder (\textit{AMSTenc}) and a mask-aware spatio-temporal decoder (\textit{MASTdec}). 
Firstly, the incomplete traffic data is fed into the encoder to generate an adaptive representation, which is subsequently transmitted to the decoder.
The decoder stacks multiple blocks incorporating a mask-aware spatio-temporal attention mechanism (\textit{MASTatt}) and attention-based spatio-temporal aggregation to capture inherent dependencies within incomplete data.
Finally, the missing values are imputed utilizing the information from correlated observations.
The specific details are discussed in the following parts.

\subsection{Adaptive Mask Spatio-temporal Encoder}
In this part, we discuss how to accurately represent incomplete traffic data without resorting to pre-filling techniques.
We claim that the incomplete traffic data is decomposed into a feature matrix $\mathbf{X}\in\mathbb{R}^{N\times T\times C}$, a mask matrix $\mathbf{M}\in\{0,1\}^{N\times T}$, and a missing matrix $\mathbf{Z}\in \mathbb{R}^{N\times T\times C}$.
$\mathbf{X}$ records the observed values, 
$\mathbf{M}$ records the missing positions,
and 
$\mathbf{Z}$ represents the missing values.
We propose an adaptive mask spatio-temporal encoder (\textit{AMSTenc}), which can adaptively represent missing values without initializing missing values with \emph{zero}.

First, we encode the feature matrix $\mathbf{X}$ to obtain the representations of observed features $\mathbf{X}_o\in\mathbb{R}^{N\times T\times d}$ through an observation embedding layer, which is denoted as follows: 
\begin{equation}
\label{observation_embedding}
    \mathbf{X}_o=\mathbf{X}\cdot \mathbf{W}_o+\mathbf{b}_o,
\end{equation}
where $\mathbf{W}_o\in\mathbb{R}^{C\times d}$, $\mathbf{b}_o\in\mathbb{R}^d$ are trainable parameters, $d$ is the size of hidden dimension.
Meanwhile, we use a learnable mask embedding layer to acquire representations $\mathbf{Z}_u\in\mathbb{R}^{N\times T\times d}$ of the missing matrix $\mathbf{Z}$.
Then, we extract the representations of the observed segment in $\mathbf{X}_o$ and the learnable representations of the missing segment in $\mathbf{Z}_u$.
We combine the extracted representations to yield the incomplete data representations $\mathbf{X}_p\in\mathbb{R}^{N\times T\times d}$:
\begin{equation}
\label{incomplete_embedding}
    \mathbf{X}_p=\mathbf{X}_o\odot \mathbf{M}+\mathbf{Z}_u\odot (\mathbf{1}-\mathbf{M}),
\end{equation}
where we broadcast $\mathbf{M}$ to the same dimension as $\mathbf{X}_o$, $\odot$ is the Hadamard product.
Next, we integrate a temporal positional embedding layer to add temporal sequence information.
Note that the temporal positional embedding is a learnable embedding rather than the sinusoidal embedding~\cite{shao2022step}.
Finally, we obtain the latent representations $\mathbf{H}\in\mathbb{R}^{N\times T\times d}$.

\subsection{Mask-Aware Spatio-temporal Decoder}
In this part, we discuss how to effectively capture inherent spatio-temporal dependencies within incomplete data.
Previous methods~\cite{yu2017stgcn,li2017dcrnn,wu2019gwnet,lan2022dstagnn,shao2022d2stgnn} rely on complete data to capture spatio-temporal dependencies, making them unsuitable for scenarios with missing data.
Spatio-temporal imputation methods such as GRIN~\cite{andrea2022grin} and GA-GAN~\cite{xu2022gagan} utilize pre-filled data to capture spatio-temporal correlations, resulting in over-smoothing at dynamic missing positions.
To address these limitations, we propose a mask-aware spatio-temporal decoder (\textit{MASTdec}) that stacks multiple blocks to capture intrinsic spatio-temporal dependencies within the latent representations $\mathbf{H}$ obtained through the \textit{AMSTenc}.
In each block of \textit{MASTdec}, we adopt a mask-aware spatio-temporal attention mechanism to assess the importance of distinct nodes.
Then, we utilize a spatial convolution based on this attention mechanism to enable the comprehensive aggregation of spatio-temporal information from diverse nodes. 
Moreover, we incorporate a multi-scale temporal convolution to update the information attributed to each  node.

\subsubsection{Mask-Aware Spatio-temporal Attention}
\ 
\newline
We design a mask-aware spatio-temporal attention mechanism (\textit{MASTatt}) to capture inherent correlations within incomplete data.
First, we utilize a multi-head self-attention~\cite{vaswani2017attention} incorporating a mask-aware mechanism to capture intricate temporal dependencies between nodes.
For the multi-head attention of $m$ heads, we define the following variables:
\begin{equation}
\label{qkv}
  \mathbf{Q}^{(h)}\triangleq \mathbf{H}\mathbf{W}_q^{(h)}, \mathbf{K}^{(h)}\triangleq \mathbf{H}\mathbf{W}_k^{(h)}, \mathbf{V}^{(h)}\triangleq \mathbf{H}\mathbf{W}_v^{(h)},
\end{equation}
\begin{equation}
\label{attention}
    \mathbf{T}_{att}=[\mathbf{T}_{att}^{(1)},\mathbf{T}_{att}^{(2)},\cdots,\mathbf{T}_{att}^{(m)}], \text{where } \mathbf{T}_{att}^{(h)}=\frac{\mathbf{Q}^{(h)}\mathbf{K}^{(h)^\mathrm{T}}}{\sqrt{d_h}},
\end{equation}
\begin{equation}
\label{residualatt}
  \mathbf{A}^{(l)}=\mathbf{T}_{att}+\mathbf{A}^{(l-1)},
\end{equation}
where $\mathbf{W}_q^{(h)}, \mathbf{W}_k^{(h)}, \mathbf{W}_v^{(h)}\in\mathbb{R}^{d\times d_h}$ are the $h$-th head parameters, $\mathbf{Q}^{(h)}, \mathbf{K}^{(h)}, \mathbf{V}^{(h)}\in\mathbb{R}^{N\times T\times d_h}, h=1,2,\cdots,m$, and $d_h$ represents the size of the hidden dimension.
Then, the attention score 
$\mathbf{T}_{att}\in\mathbb{R}^{N\times m\times T\times T}$ is derived by Eq (\ref{attention}).
We employ the attention residual connection to augment the interconnection among the mask temporal attention across multiple spatio-temporal blocks on Eq (\ref{residualatt}), where $\mathbf{A}^{(l)}\in\mathbb{R}^{N\times m\times T\times T}$ represents the temporal attention of the $l$-th spatio-temporal block, $\mathbf{A}^{(0)}=0$. 
To ignore the impact of missing values on observations, we multiply the mask matrix $\mathbf{M}$ to get the mask-aware attention context $\mathbf{C}\in\mathbb{R}^{N\times m
\times T\times d_h}$:
\begin{equation}
    \mathbf{C}=\mathrm{Softmax}(\mathbf{M}\odot\mathbf{A}^{(l)})\mathbf{V}^{(l)},
\end{equation}
\begin{equation}
\label{hmatt}
  \mathbf{H}_{matt}= \mathrm{LN}((\mathbf{C}\cdot\mathbf{W_c}+\mathbf{b_c})+\mathbf{H}).
\end{equation}
We reshape the $\mathbf{C}$, and subsequently feed to a linear layer and a residual connection with $\mathbf{H}$. 
Then the output $\mathbf{H}_{matt}\in\mathbb{R}^{N\times T\times d}$ is obtained through a normalization layer, computed as Eq (\ref{hmatt}), where $\rm LN(\cdot)$ represents the layer normalization, $\mathbf{W_c}\in\mathbb{R}^{m*d_h\times d}, \mathbf{b_c}\in\mathbb{R}^{d}$ are trainable parameters.
After that, 
we utilize a 1D convolution to aggregate the temporal representations of $\mathbf{H}_{matt}$, and then add spatial position information to obtain a high-dimensional latent representation $\mathbf{H}^{\prime}_{matt}\in\mathbb{R}^{N\times F}$ of each node, where $F$ is the dimension of spatial node embedding.
We project $\mathbf{H}^{\prime}_{matt}$ into two branches, and then calculate the mask-aware spatial attention as follows:
\begin{equation}
    \mathbf{Q}^{\prime(h)}\triangleq \mathbf{H}^{\prime}_{matt}\mathbf{W}_q^{\prime(h)},\quad \mathbf{K}^{\prime(h)}\triangleq \mathbf{H}^{\prime}_{matt}\mathbf{W}_k^{\prime(h)},
\end{equation}
\begin{equation}
    \mathbf{S}_{att}^{(h)}=\mathrm{Softmax}(\frac{\mathbf{Q}^{\prime(h)}{\mathbf{K}^{\prime(h)}}^\mathrm{T}}{\sqrt{d_h}}),
\end{equation}
\begin{equation}
\label{satt}
    \mathbf{S}_{att}=[\mathbf{S}_{att}^{(1)},\mathbf{S}_{att}^{(2)},\cdots,\mathbf{S}_{att}^{(m)}],
\end{equation}
where $\mathbf{W}_q^{\prime(h)}, \mathbf{W}_k^{\prime(h)}\in\mathbb{R}^{F\times d_h}$ are the $h$-th head parameters, and $\mathbf{Q}^{\prime(h)},\mathbf{K}^{\prime(h)}\in\mathbb{R}^{N\times d_h}, h=1,2,\cdots,m$.
Finally, we concatenate the spatial attention of $m$ heads to represent the mask-aware spatio-temporal attention $\mathbf{S}_{att}\in\mathbb{R}^{m\times N\times N}$.


\subsubsection{Attention-based Spatio-temporal Aggregation}
\
\newline
In order to further aggregate spatio-temporal features of neighboring nodes, we design an attention-based spatio-temporal aggregation module.
Unlike previous methods~\cite{yu2017stgcn,li2017dcrnn} directly aggregate the spatio-temporal features of neighboring nodes based on traffic road network in graph convolution, the mask-aware attention strategy aims to consciously adjust the aggregation.
To be specific, the information aggregated during graph convolution is the adaptive mask spatio-temporal representation $\mathbf{H}$ and the mask-aware spatio-temporal attention $\mathbf{S}_{att}$ is used as weights. 
After that, the multi-scale temporal information is aggregated using multiple temporal convolutions with diverse kernel sizes.

Regarding the graph convolution, we adopt Chebshev polynomial approximation~\cite{simonovsky2017chebshev} to learn structure-aware node features.
The laplacian matrix for Chebshev polynomial is defined as $\mathbf{L}=\mathbf{D}-\mathbf{A}$, and its normalized form is $\Tilde{\mathbf{L}}=\frac{2}{\lambda_{max}}(\mathbf{D}-\mathbf{A})-\mathbf{I}_N\in\mathbb{R}^{N\times N}$, where $\mathbf{A}$ is the adjacency matrix, $\mathbf{I}_N$ is a unit matrix, $\mathbf{D}\in\mathbb{R}^{N\times N}$ is the degree matrix,  $\mathbf{D}_{ii}=\sum_j \mathbf{A}_{ij}$, and $\lambda_{max}$ is the maximum eigenvalue of the Laplacian matrix $\mathbf{L}$.

Then, we utilize the $K$-th order Chebshev polynomials $T_k$ to aggregate the information between nodes, which is calculated as follows: 
\begin{equation}
\label{graph_conv}
    \textsl{g}_\theta *_Gx=\textsl{g}_\theta(\mathbf{L})x=\sum\limits_{k=0}^{K-1}\theta_k (T_k(\Tilde{\mathbf{L}})\odot \mathbf{S}_{att}^{(k)})x.
\end{equation}
Here, we let the graph signal at each time is $x=\mathbf{x}_t\in\mathbb{R}^{N\times C}$. 
$\textsl{g}_\theta$ denotes the approximate convolution kernel, $*_G$ denotes the graph convolution operation, and $\theta\in\mathbb{R}^K$ is the convolution kernel parameter.
Note that we accompany $T_k(\Tilde{\mathbf{L}})$ with the mask-aware spatio-temporal attention $\mathbf{S}_{att}^{(k)}$, which can dynamically adapt the information aggregation in the context of incomplete data.

After the mask-aware spatial convolution updates the spatio-temporal features of each node, we subsequently proceed to update the information corresponding to each time point.
We utilize the gated temporal convolution layer~\cite{wu2019gwnet,lan2022dstagnn} to aggregate temporal features, thereby propagating information from observed time points to missing time points.

We denote the output of Chebshev graph convolution as $\mathbf{E}^{(l)}\in\mathbb{R}^{N\times T\times d}$ in the $l$-th block.
To simplify the notation, we drop the superscript $l$.
Then, $\mathbf{E}$ is fed into $k$ gated temporal convolution blocks, where each block applies a 1D convolution kernel $\mathrm{\Gamma}\in\mathbb{R}^{1\times K \times d\times 2d}$ with a size of $1 \times K$, followed by two branches involving \emph{sigmoid} and \emph{tanh} operations.
The gated temporal convolution operation can be defined as:
\begin{equation}
    \mathrm{\Gamma} *_\gamma \mathbf{E}=\phi(\mathbf{E})\odot\sigma(\mathbf{E}),
\end{equation}
where $*_\gamma$ is gated temporal convolution operator, $\phi(\cdot)$ is the \emph{tanh} function, and $\sigma(\cdot)$ is the \emph{sigmoid} function.
The outputs of multiple gated temporal convolutions are concatenated, followed by an add operation with $\mathbf{E}$, and then passed through a ReLU activation function.
The output $\mathbf{E}_{out}\in\mathbb{R}^{N\times T\times d}$, is derived by Eq (\ref{gatedconv}), where $\mathrm{\Gamma}_1,\cdots,\mathrm{\Gamma}_k$ are convolution kernels with size $K_1,\cdots,K_k$, $||$ is concatenation.
\begin{equation}
\label{gatedconv}
    \mathbf{E}_{out}=\mathrm{ReLU}((\mathrm{\Gamma}_1 *_\gamma \mathbf{E}||\mathrm{\Gamma}_2 *_\gamma \mathbf{E}\cdots||\mathrm{\Gamma}_k *_\gamma \mathbf{E})+\mathbf{E}),
\end{equation}
\begin{equation}
\label{multi_scale_temporal_conv}
    \mathbf{H}_{out}=\mathrm{LN}(\mathrm{ReLU}(\mathbf{E}_{out}||\mathbf{H}))).
\end{equation}
Finally, to alleviate gradient dispersion, $\mathbf{E}_{out}$ is combined with $\mathbf{H}$ through a residual connection, a ReLU activation function, and a layer normalization to obtain the output $\mathbf{H}_{out}\in\mathbb{R}^{N\times T\times d}$.

\subsection{Projection Layer}
Assume the \textit{MASTdec} stacks $L$ layers of spatio-temporal blocks, we concatenate the output from each block.
Then, we utilize a two-layer fully connected network, i.e., $\mathrm{FC}$, as a output layer and apply the concatenated output to acquire imputation results $\hat{\mathbf{X}}\in\mathbb{R}^{N\times T \times C}$:
\begin{equation}
\label{projection}
    \hat{\mathbf{X}}=\mathrm{FC}(\sum\limits_{i=1}^{L}\mathbf{H}_{out}^{(i)}).
\end{equation}


\subsection{Training Strategy}
Given the ground truth $\Tilde{\mathbf{X}}\in \mathbb{R}^{N\times \tau\times C}$ on missing positions over $\tau$ slices, we optimize our model using L1 loss:
\begin{equation}
\label{loss_function}
    \mathcal{L}(\hat{\mathbf{X}},\Tilde{\mathbf{X}}, \mathbf{M})=\frac{1}{\tau}\sum\limits_{t=1}^\tau\frac{\frac{1}{N}\sum\limits_{i=1}^N \Bar{m}_t^i \cdot \frac{1}{C}\sum\limits_{j=1}^C|\hat{x}_{t,j}^i-\Tilde{x}_{t,j}^i|}{\sum\limits_{i=1}^N\Bar{m}_t^i},
\end{equation}
where $\hat{x}_{t,j}^i, \Tilde{x}_{t,j}^i$ are elements of $\hat{\mathbf{X}}, \Tilde{\mathbf{X}}$, respectively. $\Bar{m}_t^i$ is the logical binary complement of $m_t^i$. 
The training pipeline is shown in Algorithm~\ref{algorithm}.

\begin{algorithm}[H]
\SetAlgoLined
\KwIn{Incomplete traffic data: a feature matrix $\mathbf{X}$, a mask matrix $\mathbf{M}$ and a missing matrix $\mathbf{Z}$; The ground-truth of incomplete traffic data $\Tilde{\mathbf{X}}$}
\KwOut{The model parameter of optimized MagiNet $\theta$}
Random initialize $\theta$\;
\For{$e$ in $range(0,epochs,1)$}{
    Obtain the observation embedding $\mathbf{X}_o$ by Eq (\ref{observation_embedding})\;
    Obtain the learnable missing embedding $\mathbf{Z}_u$ and then yield the representation of incomplete traffic data $\mathbf{X}_p$ by Eq (\ref{incomplete_embedding})\;
    Add the temporal positional embedding and then obtain the output $\mathbf{H}$ of \textit{AMSTenc}\;
    \tcp{there is $L$ spatio-temporal blocks}
    \For{$l=1$ to $L$}  
    {
    Calculate the mask-aware spatio-temporal attention $\mathbf{T}_{att}$ and $\mathbf{S}_{att}$ by Eq (\ref{qkv}) to Eq (\ref{satt})\;
    Carry out the attention-based spatio-temporal aggregation by Eq (\ref{graph_conv}) to Eq (\ref{multi_scale_temporal_conv})\;
    Output the $\mathbf{H}^l_{out}$ of each block\;
    }
    Concatenate each block's output and utilize a projection layer to obtain the imputation results $\hat{\mathbf{X}}$ by Eq (\ref{projection})\;
    Compute the training loss $\mathcal{L}$ by Eq (\ref{loss_function})\;
    Use $\nabla_\theta\mathcal{L}$ to update $\theta$ with Adam optimizer\; 
}
return $\theta$
 \caption{The Training Process of MagiNet.}
 \label{algorithm}
\end{algorithm}

\section{Experiments}
In this section, we conduct extensive experiments to verify the effectiveness of our model, MagiNet. Specifically, the following research questions are answered:

\begin{itemize}[leftmargin=*]
    \item \textbf{RQ1:} How does MagiNet perform compared to other baselines on different datasets?
    \item \textbf{RQ2:} What is the individual contribution of each component in MagiNet to its imputation performance?
    \item \textbf{RQ3:} How does MagiNet perform with respect to different missing ratios?
    \item \textbf{RQ4:} What is the impact of each major hyperparameter on MagiNet's performance?
\end{itemize}

\subsection{Experimental Setup}

\subsubsection{Datasets.}
We evaluate our model MagiNet on five real-world traffic datasets: METR-LA, Seattle, Chengdu, Shenzhen, PEMS-BAY~\cite{lu2022spatio,xu2022gagan}. 
\begin{itemize}[leftmargin=*]
    \item \textbf{METR-LA}: The traffic dataset is collected from loop-detectors located on the Los Angeles County road network. Specifically, METR-LA contains 4 months of data recorded by 207 traffic sensors from Mar 1st, 2012 to June 30th, 2012. The data collection interval is 5 minutes, and the total number of timesteps is 34,272.
    \item \textbf{Seattle}: The traffic dataset is collected in the Seattle area from 323 detectors throughout 2015. The data collection interval is 5 minutes, and the total number of timesteps is 105,120.
    \item \textbf{Chengdu}: Traffic index dataset of Chengdu, China, is provided by Didi Chuxing GAIA Initiative. Chengdu contains data from January to April 2018, with 524 roads in the core urban area of Chengdu. The data collection interval is 5 minutes, and the total number of timesteps is 17,280.
    \item \textbf{Shenzhen}: Traffic index dataset of Shenzhen, China, is provided by Didi Chuxing GAIA Initiative. Shenzhen contains data from January to April 2018, with 627 roads in downtown Shenzhen. The data collection interval is 5 minutes, and the total number of timesteps is 17,280.
    \item \textbf{PEMS-BAY}: The traffic dataset is collected from California Transportation Agencies (CalTrans) Performance Measurement System (PeMS). Specifically, PEMS-BAY contains 6 months of data recorded by 325 traffic sensors in the Bay Area from January 1st, 2017 to June 30th, 2017. The data collection interval is 5 minutes, and the total number of timesteps is 52,116.
\end{itemize}
Consistent with the experimental settings of previous works~\cite{shao2022step,lu2020staggcn,lu2022spatio}, we adopt the following data division strategies: 
For METR-LA, Seattle, and PEMS-BAY, we sample the datasets every 12 time steps. 
About 70\% of the data is used for training, 20\% for validation, and the remaining 10\% for testing. 
For Chengdu and Shenzhen, we sample the datasets every 6 time steps.
About 60\% of the data is used for training, 20\% for validation, and the remaining 20\% for testing.
The statistical information is summarized in Table~\ref{tab:datasets}.

\begin{table}[htbp]
\caption{Statistics of traffic datasets.}
\label{tab:datasets}
\resizebox{\columnwidth}{!}{%
\begin{tabular}{@{}c|c|c|c|c|c|c|c|c|c@{}}
\toprule
Datasets & Nodes & Edges & Timesteps & Interval & \#Train & \#Valid & \#Test & Mean   & Std    \\ \midrule
METR-LA  & 207   & 1,722 & 34,272    & 5 min    & 1,999   & 285     & 571    & 54.274 & 19.664 \\
Seattle  & 323   & 1,001 & 105,120   & 5 min    & 6,132   & 876     & 1,752  & 56.854 & 12.580 \\
Chengdu  & 524   & 1,120 & 17,280    & 10 min   & 1,728   & 576     & 576    & 29.616 & 9.713  \\
Shenzhen & 627   & 4,845 & 17,280    & 10 min   & 1,728   & 576     & 576    & 31.066 & 11.126 \\
PEMS-BAY & 325   & 2,694 & 52,116    & 5 min    & 3,024   & 432     & 864    & 62.722 & 9.455  \\ \bottomrule
\end{tabular}%
}
\end{table}

\subsubsection{Baselines.}
We compare the proposed model, MagiNet, with the following baselines.
Mean, KNN~\cite{song2015knn}, MissForest~\cite{fancyimpute}, MF~\cite{kong2013matrixfactorization}, and MICE~\cite{white2011mice} are traditional statistical learning-based or machine learning-based methods.  
To compare the capability of capturing spatio-temporal dependencies within incomplete traffic data, we select five widely used traffic forecasting methods that incorporate pre-filling techniques to address the imputation task: STGCN~\cite{yu2017stgcn}, DCRNN~\cite{li2017dcrnn}, GWNet~\cite{wu2019gwnet}, DSTAGNN~\cite{lan2022dstagnn}, D2STGNN~\cite{shao2022d2stgnn}.
Last but not least, we compare eight common missing data imputation baselines:
MISGAN~\cite{li2019misgan}, 
rGAIN~\cite{yoon2018gain}, 
McFlow~\cite{richardson2020mcflow},
OT~\cite{muzellec2020ot},
BRITS~\cite{cao2018brits},
GA-GAN~\cite{xu2022gagan}, GRIN~\cite{andrea2022grin}, PriSTI~\cite{liu2023pristi}.

\begin{itemize}[leftmargin=*]
    \item Statistical learning-based or machine learning-based baselines
    \begin{itemize}
        \item \textbf{Mean}: Mean imputation, which imputes the missing values for each node using the average of the observed values.
        \item \textbf{KNN}~\cite{song2015knn}: We calculate the Euclidean distance between each node and select the average of the observations of $k$-nearest neighbor nodes to impute the missing values.
        \item \textbf{MissForest}~\cite{fancyimpute}: A tree-based imputation algorithm using the MissForest strategy.
        \item \textbf{MF}~\cite{kong2013matrixfactorization}: A method of low rank Matrix decomposition.
        \item \textbf{MICE}~\cite{white2011mice}: Impute the missing values using the Multivariate Iterative Chained Equations and multiple imputations.
    \end{itemize}
    \item Complete traffic data modeling baselines (with pre-filling techniques for imputation task)
    \begin{itemize}
        \item \textbf{STGCN}~\cite{yu2017stgcn}: Spatio-temporal graph convolution network, which integrates the graph convolution with 1D temporal convolution units.
        \item \textbf{DCRNN}~\cite{li2017dcrnn}: Diffusion convolutional recurrent neural network, which integrates the graph convolution with recurrent units.
        \item \textbf{GWNet}~\cite{wu2019gwnet}: Graph wavenet, which combines the gated TCN and GCN to jointly capture spatio-temporal dependencies.
        \item \textbf{DSTAGNN}~\cite{lan2022dstagnn}: Dynamic spatio-temporal aware graph neural network, which designs an improved multi-head attention mechanism and gated convolution to capture spatio-temporal correlations.
        \item \textbf{D2STGNN}~\cite{shao2022d2stgnn}: Decoupled dynamic spatio-temporal graph neural network, which separates the diffusion and inherent information to capture spatio-temporal dependencies.
    \end{itemize}
    \item Incomplete traffic data imputation baselines
    \begin{itemize}
        \item \textbf{MISGAN}~\cite{li2019misgan}: A GAN-based imputation framework tailored to learn from complex high-dimensional incomplete data.
        \item \textbf{rGAIN}~\cite{yoon2018gain}: A GAN-based method with a bidirectional recurrent encoder-decoder.
        \item \textbf{McFlow}~\cite{richardson2020mcflow}: A deep framework for imputation that leverages normalizing flow generative models and Monte Carlo sampling.
        \item \textbf{OT}~\cite{muzellec2020ot}: A deep neural distribution matching method based on optimal transport.
        \item \textbf{BRITS}~\cite{cao2018brits}: A bidirectional recurrent neural network for missing value imputation in time series data.
        \item \textbf{GA-GAN}~\cite{xu2022gagan}: Graph aggregate generative adversarial network, which combines GraphSAGE~\cite{hamilton2017graphsage} and GAN to impute missing values.
        \item \textbf{GRIN}~\cite{andrea2022grin}: Graph recurrent imputation network, which combines recurrent neural network and message-passing neural network.
        \item \textbf{PriSTI}~\cite{liu2023pristi}: A conditional diffusion model for spatio-temporal data imputation, which combines MPNN and conditional diffusion model.
    \end{itemize}
\end{itemize}

\subsubsection{Metrics.}
In order to verify the performance of our model, we employ the following two metrics to evaluate the imputation of missing values, including Root Mean Square Error (RMSE) and Mean Absolute Percentage Error (MAPE). The formulas are as follows:
\begin{equation}
    \mathrm{RMSE}(y,\hat{y})=\sqrt{\frac{\sum\limits_{i=1}^{N}(\hat{y}_i-y_i)^2m_i}{\sum\limits_{i=1}^{N}m_i}}, 
    \mathrm{MAPE}(y,\hat{y})=100\%\times\frac{\sum\limits_{i=1}^{N}\bigg| \frac{\hat{y}_i-y_i}{y_i} \bigg|m_i}{\sum\limits_{i=1}^{N}m_i},
\end{equation}
where $y$ is the ground truth, $\hat{y}$ is the imputed value, and $m_i$ is an element of mask matrix $\mathbf{M}$.

\subsubsection{Implementation.}
Following the previous works~\cite{xu2022gagan,andrea2022grin}, we focus on Completely Missing At Random (MCAR)~\cite{little2019statistical} to simulate the incomplete traffic data. It should be noted that we do not limit other missing patterns.
We randomly mask out 50\% of the available data.
In-depth analysis for different missing ratio are conducted in section~\ref{sensitivity}.
The number of attention heads in mask-aware spatio-temporal attention is set to 3. 
Regarding the multi-scale temporal convolution, we use 3 convolution kernels ($K$=3, 5, 7) in METR-LA, and Seattle and PEMS-BAY.
We use 2 convolution kernels ($K$=3, 5) in Chengdu and Shenzhen. 
The learning rate is searched from 10$^{-9}$ to 10$^{-2}$ with a step size 10 and the Adam optimizer is used.
We conduct a grid search on each dataset for other major hyperparameters, which is discussed in section~\ref{hyperparameters}.
The model architecture of each baseline is consistent with its original paper. For fairness, we also searched for the optimal hyperparameters of the baselines and selected the best results for comparison.
Our implementation is based on torch1.8.0. All experiments are implemented on a server with 2 CPUs (Intel Xeon Silver 4116) and 4 GPUs (NVIDIA GTX 3090). 

\subsection{Overall Performance (RQ1)}

\begin{table}[!t]
\caption{Overall performance on five real-world traffic datasets. We simulate the missing data with a missing ratio $r$=0.5 following the Missing Completely at Random (MCAR) pattern. 
We run the experiments five times and show the average results.
In each column, the \textbf{best result} is highlighted in \textbf{bold} and the \underline{second best} is \underline{underlined}. The marker $\bm{^*}$ indicates that the improvement is statistically significant compared with the best baseline (t-test with p-value $<$ 0.05).}
\label{tab:overallperformance}
\resizebox{\columnwidth}{!}{%
\begin{tabular}{@{}ccccccccccc@{}}
\toprule
\multirow{2}{*}{Methods} & \multicolumn{2}{c}{METR-LA}     & \multicolumn{2}{c}{Seattle}     & \multicolumn{2}{c}{Chengdu}     & \multicolumn{2}{c}{Shenzhen}    & \multicolumn{2}{c}{PEMS-BAY}  \\ \cmidrule(l){2-11} 
                         & RMSE           & MAPE           & RMSE           & MAPE           & RMSE           & MAPE           & RMSE           & MAPE           & RMSE          & MAPE          \\ \midrule
Mean                     & 12.69          & 18.16          & 7.44           & 18.76          & 6.42           & 22.10          & 7.07           & 21.58          & 5.12          & 8.68          \\
KNN~\cite{song2015knn}                      & 7.43           & 10.31          & 5.88           & 10.08          & 4.60           & 13.57          & 4.53           & 12.03          & 3.46          & 4.01          \\
MissForest~\cite{fancyimpute}               & 6.32           & 8.47           & 4.75           & 8.44           & 4.16           & 12.10          & 4.24           & 10.69          & 2.76          & 3.37          \\
MF~\cite{kong2013matrixfactorization}                       & 5.75           & 8.42           & 4.53           & 7.94           & 4.30           & 13.04          & 4.65           & 12.67          & 2.61          & 3.23          \\
MICE~\cite{white2011mice}                     & 6.19           & 8.97           & 4.75           & 8.35           & 3.96           & 11.27          & 3.76           & 9.65           & 2.50          & 3.14          \\ \midrule
STGCN~\cite{yu2017stgcn}& 7.00           & 8.95           & 4.01           & 6.45           & 3.48           & 10.60          & 3.64           & 10.52          & 3.28          & 3.97          \\
DCRNN~\cite{li2017dcrnn}                    & 14.75          & 7.25           & 3.47           & {\ul 5.24}     & 3.71           & 10.46          & 3.50           & 8.73           & 2.25          & 2.22          \\
GWNet~\cite{wu2019gwnet}                    & 7.50           & 7.30           & 4.13           & 6.29           & 3.78           & 10.77          & 3.60           & 9.12           & 2.22          & 2.14          \\
DSTAGNN~\cite{lan2022dstagnn}                  & 6.38           & 7.44           & 3.78           & 5.91           & {\ul 3.16}     & 9.39           & {\ul 2.94}     & {\ul 8.14}     & 1.95          & 2.03          \\
D2STGNN~\cite{shao2022d2stgnn}                  & 14.03          & 8.08           & {\ul 3.43}     & 5.33           & 3.20           & 9.55           & 2.95           & {\ul 8.14}     & {\ul 1.82}    & 2.02          \\ \midrule
MISGAN~\cite{li2019misgan}                   & 11.97          & 14.81          & 6.82           & 13.63          & 4.61           & 15.82          & 4.49           & 13.01          & 5.48          & 7.04          \\
rGAIN~\cite{yoon2018gain}                    & 10.36          & 13.38          & 6.28           & 11.60          & 3.86           & 12.16          & 3.72           & 10.68          & 4.51          & 5.69          \\
McFlow~\cite{richardson2020mcflow}                   & 9.50           & 11.71          & 5.67           & 10.55          & 3.54           & 10.90          & 3.37           & 9.19           & 4.24          & 5.16          \\
OT~\cite{muzellec2020ot}                       & 7.23           & 10.03          & 5.35           & 11.52          & 4.78           & 13.75          & 5.72           & 13.32          & 3.59          & 5.68          \\
BRITS~\cite{cao2018brits}                    & 9.24           & 11.27          & 6.15           & 12.60          & 4.26           & 14.12          & 4.23           & 12.40          & 3.00          & 3.20          \\
GA-GAN~\cite{xu2022gagan}                   & 7.15           & 8.24           & 4.11           & 7.39           & 3.51           & 10.75          & 3.26           & 8.82           & 2.10          & 2.18          \\
GRIN~\cite{andrea2022grin}                     & 6.24           & 7.93           & 4.30           & 7.56           & 4.05           & 12.85          & 3.47           & 9.85           & 2.61          & 3.03          \\
PriSTI~\cite{liu2023pristi}                   & {\ul 5.34}     & {\ul 6.39}     & 3.96           & 5.77           & 3.47           & {\ul 8.95}     & 3.30           & 8.37           & \textbf{1.80} & \textbf{1.83} \\ \midrule
MagiNet (ours)           & \bm{${4.96}$}* & \bm{$6.35$}* & \bm{$3.35$}* & \bm{$5.10$}* & \bm{$2.87$}* & \bm{$8.16$}* & \bm{$2.74$}* & \bm{$7.39$}* & $1.87$          & {\ul $1.88$}    \\ \bottomrule
\end{tabular}%
}
\end{table}

The overall performance is shown in Table~\ref{tab:overallperformance}. We make the following observations: 
(1) MagiNet outperforms other baselines on multiple datasets, yielding an average improvement of 4.31\% in RMSE and 3.72\% in MAPE, demonstrating the superiority of our method. 
(2) 
Complete traffic data modeling methods with pre-filling techniques exhibit unstable performance across different datasets and are inferior to our MagiNet. 
This deficiency is attributed to their limitations in capturing the inherent spatio-temporal dependencies within incomplete traffic data.
In contrast, MagiNet utilizes mask-aware spatio-temporal attention, adaptively adjusting aggregation coefficients based on observation information, which extracts the intrinsic dependencies and alleviates the over-smoothing effect.
Consequently, MagiNet surpasses these complete traffic data modeling methods with pre-filling techniques, demonstrating an average performance enhancement of 7.56\% and 8.87\% in RMSE and MAPE, respectively.
(3) MagiNet significantly outperforms existing missing data imputation methods, as these methods rely on pre-filling techniques to initialize missing values, introducing uncontrollable noise and resulting in error propagation.
Instead, MagiNet adopts the learnable missing encoding approach to represent missing values, avoiding noise introduction.
(4) PriSTI performs slightly better than MagiNet on PEMS-BAY dataset, potentially attributed to PriSTI's imputation of missing values through a substantial number of generation steps, making it more suitable for datasets with low variance such as PEMS-BAY. 
However, MagiNet consistently demonstrates strong performance across other datasets.
Further experiments about the effectiveness of our MagiNet are shown in section~\ref{ablation} and section~\ref{casestudy}.

\begin{table}[!t]
\caption{Ablation study of MagiNet on five datasets. We run the experiments five times and show the average results.
The marker $\bm{^*}$ indicates that the improvement is statistically significant compared with the best baseline (t-test with p-value $<$ 0.05).}
\label{tab:ablationstudy}
\resizebox{\columnwidth}{!}{%
\begin{tabular}{@{}ccccccccccc@{}}
\toprule
\multirow{2}{*}{Method} & \multicolumn{2}{c}{METR-LA}     & \multicolumn{2}{c}{Seattle}     & \multicolumn{2}{c}{Chengdu}     & \multicolumn{2}{c}{Shenzhen}    & \multicolumn{2}{c}{PEMS-BAY}   \\ \cmidrule(l){2-11} 
                        & RMSE           & MAPE           & RMSE           & MAPE           & RMSE           & MAPE           & RMSE           & MAPE           & RMSE          & MAPE           \\ \midrule
zero prefill           & 5.05           & 6.57           & 3.42           & 5.27           & 2.99           & 8.56           & 2.82           & 7.70           & 1.90          & 1.91           \\
mean prefill           & 5.08           & 6.49           & 3.40           & 5.22           & 3.11           & 9.12           & 2.80           & 7.65           & 1.89          & 1.91           \\
w/o AMSTenc             & 5.19           & 6.46           & 3.45           & 5.29           & 3.56           & 10.07          & 3.47           & 8.76           & 1.91          & 1.92           \\\midrule
w/o MASTatt             & 4.99           & 6.45           & 3.36           & 5.11           & 3.28           & 9.52           & 2.88           & 7.87           & 1.89          & 1.91           \\
w/o Graphconv             & 7.14           & 7.12           & 3.96           & 6.07           & 3.37           & 9.83           & 3.00           & 8.18           & 1.93          & 1.93           \\
w/o GTconv              & 5.06           & 6.43           & 3.40           & 5.23           & 2.97           & 8.52           & 2.79           & 7.58           & \textbf{1.88} & 1.90           \\
w/o MASTdec             & 6.94           & 7.50           & 4.01           & 6.12           & 3.37           & 9.79           & 2.99           & 8.10           & 2.02          & 2.07           \\\midrule
MagiNet                 & \textbf{4.96*} & \textbf{6.35*} & \textbf{3.35*} & \textbf{5.10*} & \textbf{2.87*} & \textbf{8.16*} & \textbf{2.74*} & \textbf{7.39*} & \textbf{1.88} & \textbf{1.89*} \\ \bottomrule
\end{tabular}%
}
\end{table}

\subsection{Ablation Study (RQ2)}
\label{ablation}
To verify the effectiveness of each component of MagiNet, we systematically modify or remove individual components to create variants of MagiNet and do grid search to optimize the performance of these variants. 
Seven types of MagiNet variants are considered. 
We set (1) \emph{zero prefill} and (2) \emph{mean prefill} to assess the performance of pre-filling missing values with zero and mean values, respectively. 
(3) We set \emph{w/o AMSTenc} to test the performance of replacing the adaptive mask spatio-temporal encoder (\emph{AMSTenc}) with a regular embedding layer. 
(4) We set \emph{w/o MASTatt} to test the performance of MagiNet without mask-aware spatio-temporal attention (\emph{MASTatt}) mechanism. 
(5) We set \emph{w/o Graphconv} to test the performance of MagiNet without attention-based graph convolution (\emph{Graphconv}).
(6) We set \emph{w/o GTconv} to test the performance of MagiNet without multi-scale gated temporal convolution (\emph{GTconv}).
(7) We set \emph{w/o MASTdec} to test the performance of replacing the mask-aware spatio-temporal decoder (\emph{MASTdec}) with a single regression layer. 

As shown in Table~\ref{tab:ablationstudy}, MagiNet outperforms other variants across five datasets. 
The superior performance of MagiNet over \emph{zero prefill} and \emph{mean prefill} indicates the positive impact of learnable missing encoding. 
Moreover, MagiNet is superior to \emph{w/o AMSTenc}, suggesting that the adaptive mask spatio-temporal encoder plays a vital role in learning the representation of incomplete traffic data. 
MagiNet surpasses \emph{w/o MASTatt}, highlighting the positive role of mask-aware spatio-temporal attention.
MagiNet also surpasses \emph{w/o Graphconv} and \emph{w/o GTconv}, which indicates that the attention-based graph convolution and multi-scale gated temporal convolution modules are also important to aggregate information.
Furthermore, MagiNet significantly outperforms \emph{w/o MASTdec}, showing the significance of capturing inherent spatio-temporal correlations.
\begin{figure}[!t]
    \centering
    \includegraphics[width=0.85\textwidth]{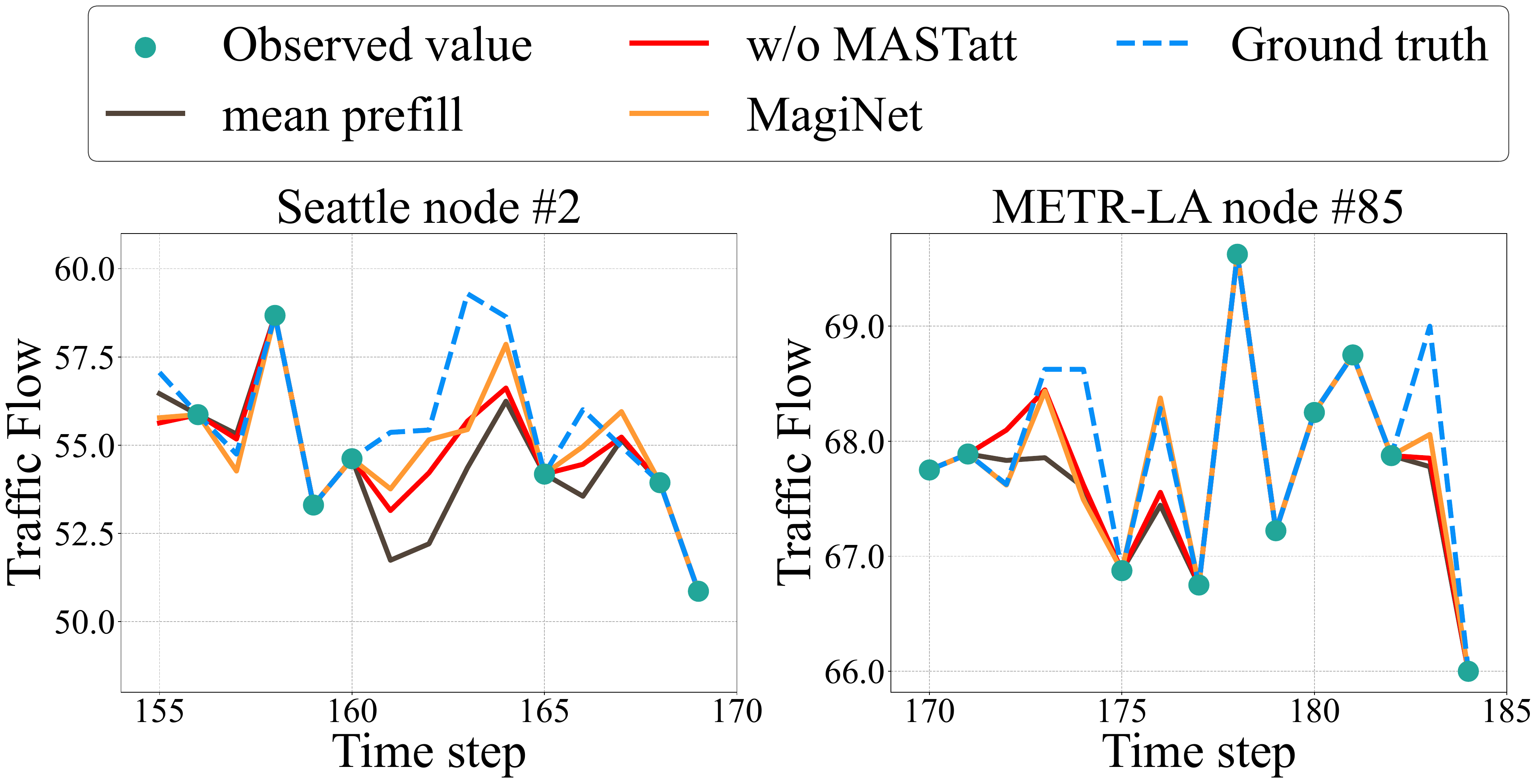}
    \caption{Comparison of imputation curves between MagiNet and its variants for two snapshots: node \#2 in Seattle and node \#85 in METR-LA.}
    \label{fig:case_study}
\end{figure}

\subsection{Case Study (RQ2)}
\label{casestudy}
To intuitively understand how MagiNet imputes the incomplete traffic data, we visualize the imputation curves between MagiNet and its variants on two datasets with high variance, Seattle and METR-LA.
Specifically, we randomly select two snapshots of imputation results: node \#2 in Seattle and node \#85 in METR-LA.
As shown in Figure~\ref{fig:case_study}, MagiNet outperforms \emph{mean prefill} and \emph{w/o MASTatt}, especially in continuous missing positions and dynamic positions, such as time step 160 to 168 in Seattle and time step 171 to 176 in METR-LA. 
The capability of MagiNet to mitigate imputation errors can be attributed to \emph{AMSTenc}, which substitutes pre-filling techniques and avoids introducing noise, and \emph{MASTatt}, designed to capture inherent spatio-temporal dependencies within incomplete traffic data. 

\subsection{Sensitivity Analysis (RQ3)}
\label{sensitivity}
\begin{figure}[!t]
    \centering
    \includegraphics[width=\textwidth]{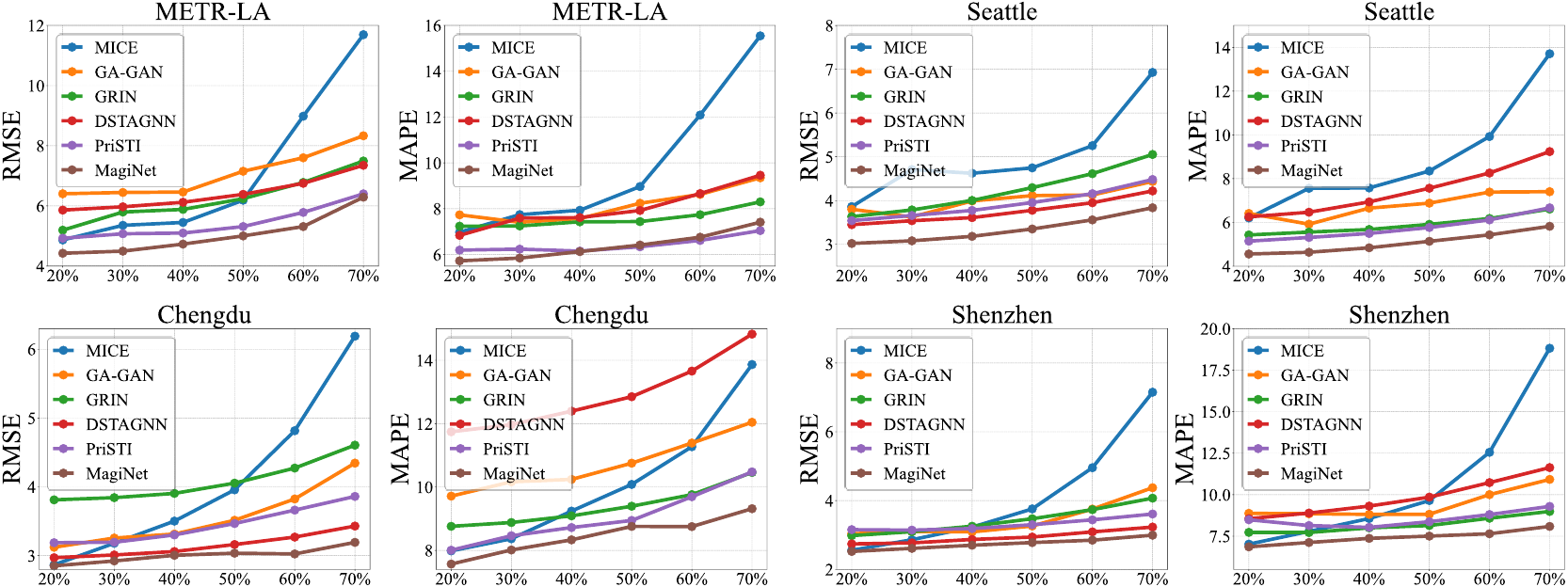}
    \caption{Sensitivity analysis on METR-LA, Seattle, Chengdu, and Shenzhen datasets with respect to different missing ratio $r$ from 20\% to 70\%.}
    \label{fig:sensitivity}
\end{figure}

We carry out an assessment of performance consistency with respect to different missing ratios.
We present the results on METR-LA, Seattle, Chengdu, and Shenzhen datasets. 
To take into account comprehensively of the sufficiency and sparsity of missing data, we select the missing ratio $r$ from 20\% to 70\%.
This is because when the missing ratio is too low, the model may simply use the values near the missing points for interpolation. Conversely, when the missing ratio is too high, there is little valuable information, resulting in rapidly deterioration in model performance. 
Then, we select five representative models from the baselines: MICE~\cite{white2011mice}, GA-GAN~\cite{xu2022gagan}, GRIN~\cite{andrea2022grin}, and DSTAGNN~\cite{lan2022dstagnn}, PriSTI~\cite{liu2023pristi} to compare with MagiNet.
The results are shown in Figure~\ref{fig:sensitivity}. 
The imputation performance degrades as the missing ratio increases, aligning with our intuition.
MagiNet consistently performs as the best, with the lowest degradation speed, suggesting that MagiNet demonstrates the capability to maintain remarkable consistency across various missing ratios.

\begin{figure}[!t]
    \centering
    \includegraphics[width=\textwidth]{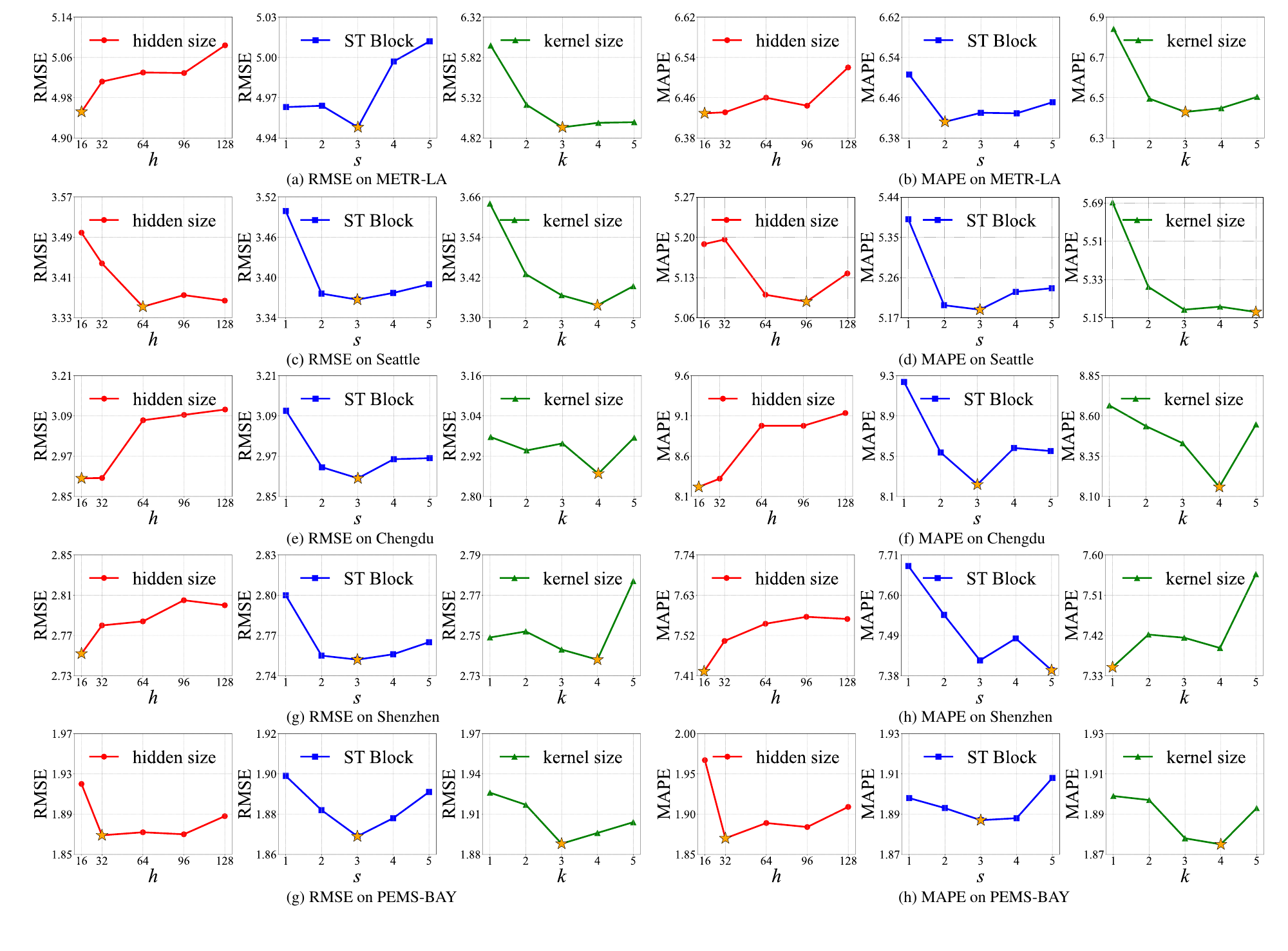}
    \caption{Hyperparameter study on three key parameters of MagiNet: hidden size $\emph{h}$, spatio-temporal blocks $\emph{s}$, mask-aware spatial kernel size $\emph{k}$.}
    \label{fig:hyperparameter_study}
\end{figure}
\subsection{Hyperparameter Study (RQ4)}
\label{hyperparameters}
We determine the hyperparameters by analyzing the experimental results as illustrated  in Figure~\ref{fig:hyperparameter_study}.
(1) We vary the size of hidden dimension of incomplete traffic data $\emph{h}$ from 16 to 128, which impacts the model's capability to capture inherent features.
Overall, the optimal performance can be achieved when $\emph{h}=16$.
Small $\emph{h}$ is insufficient for learning incomplete traffic data representation, while large $\emph{h}$ is prone to overfitting.
(2) We vary the number of spatio-temporal blocks $\emph{s}$ ranging from 1 to 5, which directly influences the extent of intrinsic spatio-temporal correlations extracted by the model. 
When $s$ is small, the spatio-temporal correlation capturing is insufficient, while $s$ is large, the model overfits some outliers, leading to a decrease in performance.
(3) We adjust the size of spatial convolution kernel $\emph{k}$ between 1 and 5, which determines the range of aggregated neighbor information.
Setting $\emph{k}=3$ or $\emph{k}=4$ often yields better performance.

\subsection{Visualization Results}
\begin{figure}[htbp]
    \centering
    \includegraphics[width=\textwidth]{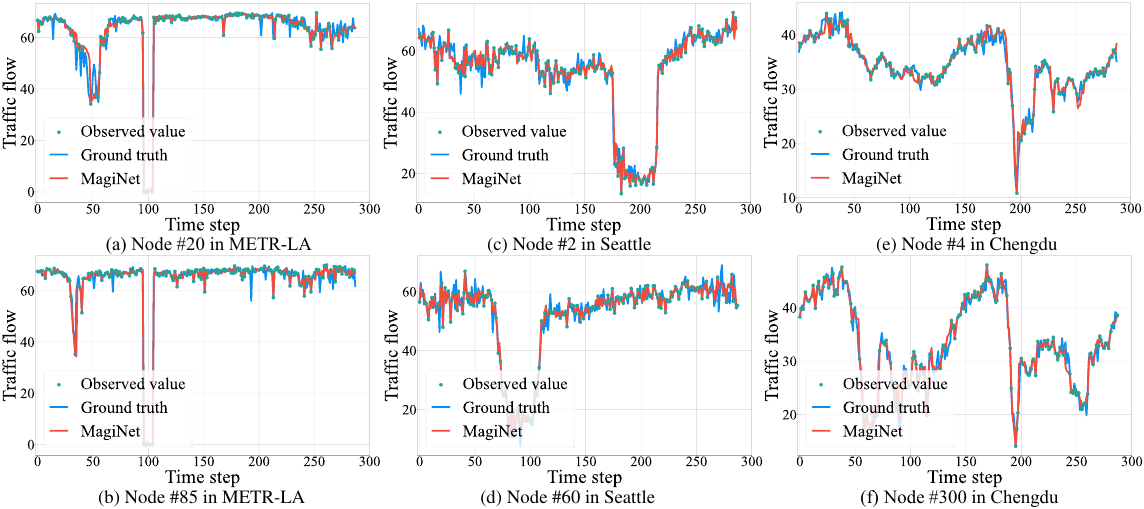}
    \caption{Imputation results of incomplete traffic
data in METR-LA, Seattle, and Chengdu.}
    \label{fig:imputation_case1}
\end{figure}

\begin{figure}[htbp]
    \centering
    \includegraphics[width=0.75\textwidth]{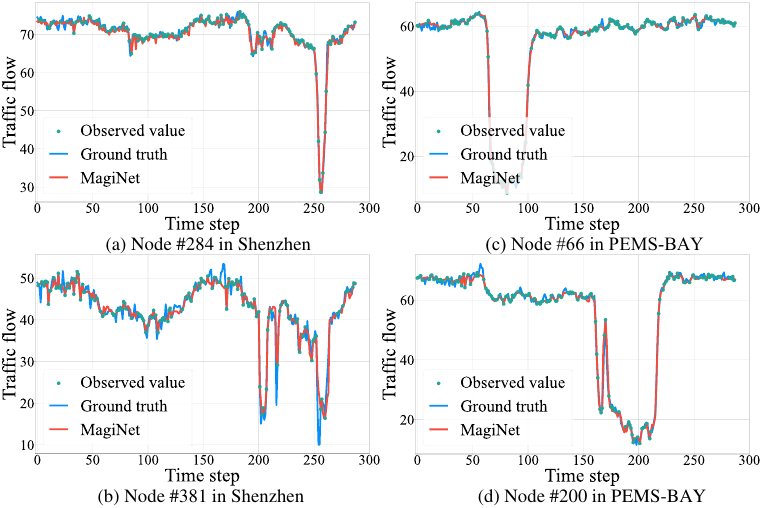}
    \caption{Imputation results of incomplete traffic
data in Shenzhen and PEMS-BAY.}
    \label{fig:imputation_case2}
\end{figure}

To further intuitively understand the imputation performance of our model MagiNet, we provide more visualization of incomplete traffic data imputation results on five datasets, as shown in Figure~\ref{fig:imputation_case1} and Figure~\ref{fig:imputation_case2}. 
We randomly select two nodes from each dataset to demonstrate the imputation and ground truth within a single day. 

\section{Conclusion}
In this paper, we propose MagiNet, a mask-aware graph imputation network to impute missing values within incomplete traffic data. 
MagiNet employs an adaptive mask spatio-temporal encoder to represent the incomplete traffic data and a mask-aware spatio-temporal decoder that stacks multiple blocks to capture inherent spatio-temporal dependencies in the presence of missing values. 
Experimental results demonstrate that our MagiNet achieves outstanding performance across five public datasets for incomplete traffic data imputation task.
In the future, we intend to extend our MagiNet to address probabilistic imputation task and explore its scalability on larger scale traffic dataset.


\bibliographystyle{ACM-Reference-Format}
\bibliography{tkdd}


\end{document}